\documentclass{article} 
\usepackage[preprint]{colm2026_conference}

\usepackage{microtype}
\usepackage{hyperref}
\usepackage{url}
\usepackage{booktabs}
\usepackage{amsmath,amssymb}
\usepackage{graphicx}
\usepackage{tcolorbox}
\usepackage{tabularx}
\tcbuselibrary{listings,breakable}
\usepackage{sidecap} 
\newcommand{\ours}{\textsc{GoodPoint}}
\newcommand{\ourdataset}{\textsc{GoodPoint-ICLR}}

\usepackage{lineno}
\usepackage{graphicx}
\usepackage{xcolor}
\usepackage{tikz}
\usetikzlibrary{positioning, calc}
\usepackage{mdframed}
\usepackage{multirow}
\usepackage{subcaption}

\definecolor{darkblue}{rgb}{0, 0, 0.5}
\hypersetup{colorlinks=true, citecolor=darkblue, linkcolor=darkblue, urlcolor=darkblue}

\title{\textsc{GoodPoint}: Learning Constructive\\Scientific Paper Feedback from Author Responses}

\author{
Jimin Mun$^{1\heartsuit}$\hspace{3mm} 
Chani Jung$^{2\heartsuit}$\thanks{Work was done while Chani was an intern at CMU.}\hspace{3mm} \textbf{Xuhui Zhou}$^1$\hspace{3mm} 
\textbf{Hyunwoo Kim}$^{3\diamondsuit}$\hspace{3mm} 
\textbf{Maarten Sap}$^{1\diamondsuit}$\\
$^1$Carnegie Mellon University\hspace{3mm}
$^2$Independent Researcher\hspace{3mm}
$^3$NVIDIA\vspace{1mm}\\
{\small $^\heartsuit$Equal contribution. $^\diamondsuit$Equal advising contribution.}\vspace{1mm}\\
\texttt{jmun@andrew.cmu.edu}
}

\begin{document}

\ifcolmsubmission
\linenumbers
\fi

\maketitle

\begin{abstract}
While LLMs hold significant potential to transform scientific research, we advocate for their use to augment and empower researchers rather than to automate research without human oversight. To this end, we study \textit{constructive feedback generation}, the task of producing targeted, actionable feedback that helps authors improve both their research and its presentation. In this work, we operationalize the effectiveness of feedback along two author-centric axes---\textit{validity} and \textit{author action}. We first curate \ourdataset{}, a dataset of 19K ICLR papers with reviewer feedback annotated along both dimensions using author responses. Building on this, we introduce \ours{}, a training recipe that leverages success signals from author responses through fine-tuning on valid and actionable feedback, together with preference optimization on both real and synthetic preference pairs.
Our evaluation on a benchmark of 1.2K ICLR papers shows that a \ours{}-trained Qwen3-8B improves the predicted success rate by 83.7\% over the base model and sets a new state-of-the-art among LLMs of similar size in feedback matching on a golden human feedback set, even surpassing Gemini-3-flash in precision. We further validate these findings through an expert human study, demonstrating that \ours{} consistently delivers higher practical value as perceived by authors.\footnote{We will release code, dataset, and trained models upon acceptance.}
\looseness=-1
\end{abstract}


\section{Introduction}
The growing use of LLMs across scientific research \citep{si2025ideation,huang2025deep,chen2025ai4research,ifargan2025autonomous,schmidgall2025agent} has spurred proposals to automate core research processes, including peer review \citep{zhu2025deepreview,zhou-etal-2024-llm,liang2024can}. Yet unchecked automation risks eroding the critical judgment of researchers and the broader scientific community, ultimately degrading the quality of science \citep{baumannstop,Wilder_2025}.

We argue that AI should instead empower researchers—particularly those who are traditionally disadvantaged \citep[junior and/or non-native English speaking researchers;][]{liao2024llms}—rather than replace human scientific judgment. A high-value opportunity lies in using LLMs to deliver \emph{constructive feedback} \citep{jefferson2002effects,smith2006peer}. This is especially consequential for authors who lack access to timely, expert feedback \citep{liao2024llms}. However, generating such feedback for scientific papers is non-trivial: it demands deep domain knowledge, which is challenging even for human experts \citep{alberts2008reviewing} and unsurprisingly difficult for LLMs as well \citep{liang2024can,steiss2024comparing}.\looseness=-1

Existing LLM-generated reviews reflect this difficulty, exhibiting weak specificity, limited actionability \citep{liang2024can}, and poor alignment with human reviewers in feedback prioritization \citep{steiss2024comparing}. Addressing these gaps requires confronting two fundamental challenges: (1) rigorously defining what constitutes constructive, effective feedback for improving scientific research, and (2) building training and evaluation frameworks that operationalize this definition. In this work, we tackle both through an author-centric lens, leveraging author responses in author--reviewer discussion as a natural signal to define, train, and evaluate feedback quality.

\begin{figure}[t]
\begin{center}
\includegraphics[width=0.85\linewidth]{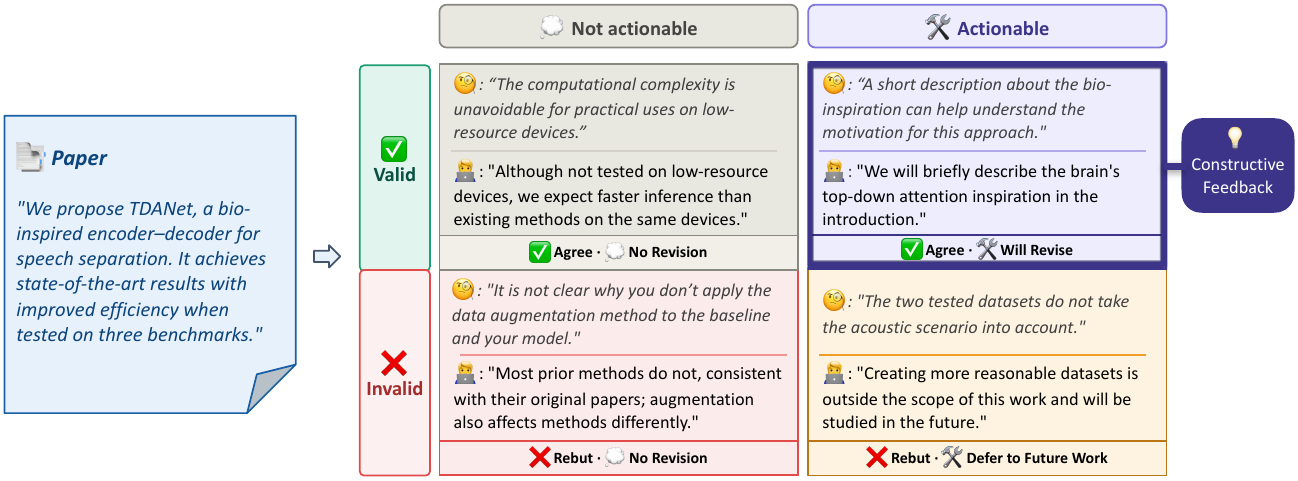}
\end{center}
\caption{Overview of the two feedback quality axes and our definition of \emph{constructive feedback}. Reviewer feedback is evaluated along two dimensions: validity---whether the raised concern is agreed upon by the authors---and actionability---whether it prompts concrete follow-up actions. We consider the feedback that is both valid and actionable as being successful, and build training and evaluation frameworks that operationalize this definition.\looseness=-1}
\label{fig:fb_quality_axes}
\end{figure}

To address the first challenge, we introduce two author-centric axes of feedback quality: validity and author action (§\ref{sec:problem-statement}). As shown in Figure~\ref{fig:fb_quality_axes}, validity captures whether the authors agree with a raised concern, as invalid feedback tends to be rebutted. Author action captures whether authors propose concrete follow-up actions in response, noting that feedback can be valid yet not immediately actionable, still prompting meaningful discussion (e.g., suggestions deferred to future work). 

Building on this definition, we operationalize constructiveness for both training and evaluation.
We curate \ourdataset{}, which contains initial manuscript submissions and author--reviewer discussions for 19,534 ICLR papers (2020--2026), with each reviewer comment labeled for validity and author action (§\ref{sec:goodpoint-iclr}). 
We then introduce \ours{}, a training recipe for constructive review generation.
We first fine-tune Qwen3-8B on successful (i.e., valid and actionable) human reviews and then further align it via DPO on paired feedback data derived from both success labels and synthetic corruptions targeting key quality dimensions (\S\ref{sec:goodpoint}). 
To evaluate LLM-generated feedback, we develop two complementary automatic evaluation frameworks---author response prediction and human consensus-based feedback evaluation---grounding on our definition of constructive feedback (\S\ref{sec:evaluation}). \looseness=-1

When evaluated on a held-out test set of 1,198 papers, \ours{}-trained Qwen3-8B achieves 83.7\% improvement in predicted success rate compared to the base model. In feedback matching against multi-reviewer consensus feedback, \ours{}-SFT improves F1 by 58.8\% over the base model and even exceeds Gemini-3-flash and GPT-5.2 in precision, suggesting that our model produces more selective critiques despite its substantially smaller scale. In an expert human study, \ours{}-DPO outperforms Qwen3-8B across validity, actionability, specificity, and helpfulness, meaningfully reducing the gap to Gemini-3-flash. This result demonstrates that our automated evaluation results translate into practical usefulness perceived by real-world authors. 
Together, these findings highlight the promise of training and evaluating LLMs grounded in an author-centric definition of constructive feedback, and underscore the potential of LLMs to augment---rather than replace---human researchers.\looseness=-1







\section{Related Works}
LLMs are increasingly integrated into scientific workflows \citep{chen2025ai4research, huang2025deep, si2025ideation}, yet fully autonomous systems risks distorting scientific judgment \citep{baumannstop, Wilder_2025}. Instead, we advocate for human-AI collaboration. Although automated peer review has emerged to alleviate the burden of reviewers \citep{zhou2024llm, liang2024can}, current LLM-generated reviews often lack critical depth and tend to over-positivity \citep{zhou2024llm, idahl-ahmadi-2025-openreviewer}. Existing approaches rely on specialized fine-tuning \citep{idahl-ahmadi-2025-openreviewer}, review standardization \citep{yu-etal-2024-automated}, or literature retrieval \citep{zhu2025deepreview}. However, unlike these systems that prioritize mimicking reviewer behavior, \ours{} focuses specifically on \emph{constructive feedback}. By leveraging author responses as an explicit training signal, we prioritize feedback that is both valid and actionable from the researcher's perspective. See Appendix~\ref{app:extended-related-works} for an extended discussion. \looseness=-1


\section{\ours: Constructive Feedback Learning Framework}
\label{sec:goodpoint}
We introduce \ours{}, a framework for generating author-centric paper feedback. Our approach builds on a dataset of ICLR 2020–2025 review discussions, from which we extract \textit{successful} feedback items — those that authors both acknowledged as valid and committed to act upon — as the core training signal. We first fine-tune an LLM on this curated set via SFT to ground the model in constructive feedback patterns. We then sharpen both content and style through two DPO preference sets: one contrasting valid versus invalid feedback, and another using targeted corruptions along five dimensions — specificity, clarity, accuracy, prioritization, and supportive tone — as rejected samples.

\subsection{\ours{} Definition}
\label{sec:problem-statement}
To operationalize our goal of generating constructive feedback, we use author responses to define a measurable signal of feedback success. Concretely, given a paper $x$ and a set of self-contained feedback items $F = \{f_1, f_2, \ldots, f_n\}$ with corresponding author responses $R = \{r_1, r_2, \ldots, r_n\}$, we define a feedback item $f_i$ as \textbf{successful} if the author acknowledges it as both \textit{valid} and \textit{actionable} (examples in Figure~\ref{fig:fb_quality_axes}). Formally, we define an indicator function:

\begin{equation}
    \mathbb{1}_{\text{success}}(f_i, r_i) =
    \begin{cases}
        1 & \text{if } \mathcal{V}(f_i, r_i) = 1 \text{ and } \mathcal{A}(f_i, r_i) = 1 \\
        0 & \text{otherwise}
    \end{cases}
\end{equation}

\noindent where $\mathcal{V}(f_i, r_i) \in \{0, 1\}$ denotes whether the author response $r_i$ acknowledges feedback $f_i$ as \textbf{valid} (i.e., the critique is factually correct and relevant to the paper), and $\mathcal{A}(f_i, r_i) \in \{0, 1\}$ denotes whether $r_i$ acknowledges $f_i$ as \textbf{actionable} (e.g., commit to a revision or defer to future work).
Our goal is to learn a feedback generation model $\mathcal{M}_\theta$ that, given 
a paper $x$, produces a feedback set $F^* = \mathcal{M}_\theta(x)$ maximizing the 
expected success rate.



\subsection{\ourdataset{}}
\label{sec:goodpoint-iclr}
We construct our dataset from ICLR submissions spanning 2020 to 2026. Papers are sourced from three outlets: Re$^2$ \citep{zhang2025re} for 2020–2023, arXiv via title matching for 2024–2025, and direct PDF submissions to OpenReview for 2026. We convert all paper into markdown format using Marker\footnote{https://github.com/datalab-to/marker}. Each submission is paired with its corresponding review thread from OpenReview. We partition the data so that ICLR 2020–2025 serves as the training, development, and test split, while ICLR 2026 is held out exclusively as a temporally separate test set. This ensures a contamination-free evaluation against all baselines, all of which have knowledge cutoffs prior to August 2025. Our dataset includes a total of 18,936 papers with 14,517 rejected papers and 4,419 accepted papers. 

We process each review thread with GPT-4.1 (\texttt{gpt-4.1-2025-04-14}), parsing it into self-contained feedback units paired with author responses. Each unit is annotated with two binary labels: \textit{validity}, whether the author agreed with or rebutted the feedback, and \textit{actionability}, whether the author's response falls into one of six action classes---\textit{will revise, defer to future work, point to existing content, no revision accept, no revision contest}, and \textit{no action other}---where the first two are considered actionable. Human verification yields 0.936 and 0.941 accuracy for validity and action parsing, with moderate and strong inter-annotator agreement \citep[PABAK $=$ 0.747 and 0.837 respectively;][]{byrt1993bias}. Full details on prompts, action examples, and parsing performance appear in Appendix~\ref{app:parsing-setup}.

\paragraph{Feedback Corruption for Preference Pair Construction}
To construct informative preference pairs for DPO, we apply targeted quality corruptions \citep{li2025alfa,geng2025the} to successful reviewer-written feedback. We take five characteristics of good feedback from previous work in education research \citep{steiss2024comparing}: specific, clear and actionable, accurate, essential, and supportive, and corrupt reviewer-written feedback in only one dimension. To ensure sufficient delta in our corrupted samples, we only select valid and actionable feedback from our dataset to corrupt. Furthermore, we validate the corruption through an LLM-as-a-judge framework and filter based on its accurate prediction of the corrupted dimension (prediction accuracy $>0.94$ for all dimensions except generic, 0.62) and scoring of both degradation and collateral preservation. See Appendix~\ref{app:feedback-corruption} for more details on feedback corruption and verification.

\begin{figure}[t]
\begin{center}
\includegraphics[width=0.85\linewidth]{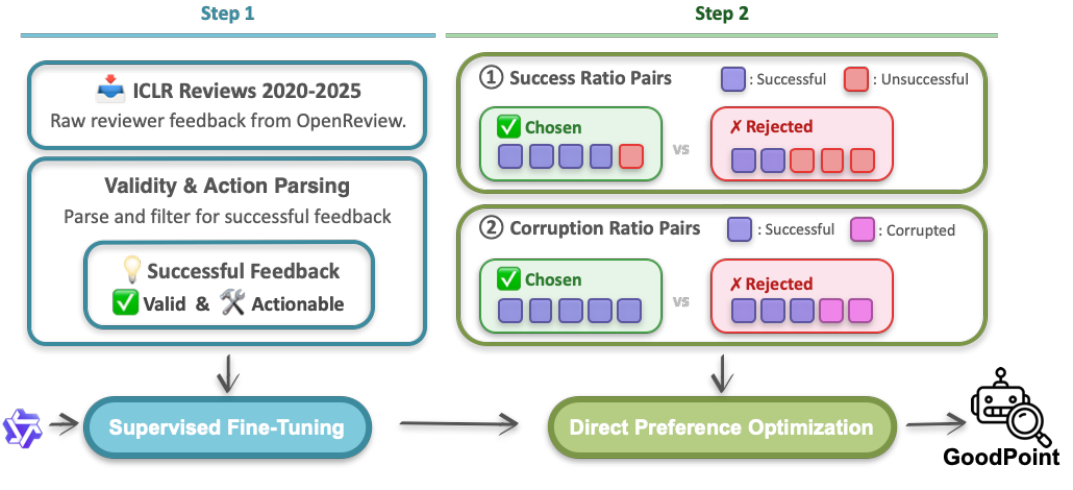}
\end{center}
\caption{Overview of \ours{} training strategy. We first fine-tune Qwen3-8B on successful (i.e., valid and actionable) feedback from ICLR 2020–2025. We then apply DPO using two types of preference pairs: (1) those contrasting varying proportions of unsuccessful items, and (2) those featuring successful items with synthetic corruptions targeting specificity, clarity, accuracy, prioritization, and supportive tone. In both cases, DPO prioritizes responses with minimal failure or corruption.
}
\label{fig:goodpoint_training}
\end{figure}

\subsection{Training}
To train a constructive feedback generation model, we use both supervised fine-tuning (SFT) and direct preference optimization (DPO). We first ground the model's feedback generation through SFT on \textit{successful} (valid and actionable) feedback. Then we use DPO to further align model to generate effective feedback in both content and style. Due to our limited computational resources, we use Qwen3-8B as our base model. 

\paragraph{\ours{}-SFT}
We fine-tune a base LLM on the valid and actionable feedback units from \ourdataset{} (6K papers), formatted as (paper, aggregated feedback) pairs. We aggregate all feedback given by a single reviewer for a particular paper for the model to learn a consistent and diverse style of feedback. This grounds the model in the patterns of constructive, author-acknowledged feedback before preference optimization.

\paragraph{\ours{}-DPO}
We further align the SFT model using two preference datasets. The first contrasts valid versus invalid feedback items drawn directly from \ourdataset{} (9K papers). The second uses our corruption pipeline to form pairs of original versus corrupted feedback, targeting the five quality dimensions described above. We construct DPO samples such that the acceptance ratio is higher in the chosen set compared to rejected at least by 2 feedback items. In this phase of training, we aggregate all the reviews for one paper for comprehensive coverage of feedback and deduplicate items across reviewers using embedding similarity. Training details and hyperparameters are provided in Appendix~\ref{app:training-setup}.

\section{Evaluations}
\label{sec:evaluation}
To robustly assess feedback quality, we evaluate feedback quality using two complementary automatic metrics and human expert annotations. First, we predict author responses to a paper–feedback pair as proxy measures of validity and actionability. Second, we construct a high-agreement reference set and measure how closely model-generated feedback matches it. Together, these metrics capture feedback quality both top-down and bottom-up, while the second additionally measures coverage of essential and prioritized feedback. We describe each protocol in turn.

\subsection{Author Response Prediction}
To assess author-focused effective feedback generation, we use author response prediction as a proxy of success as defined in \S\ref{sec:problem-statement}. Given an input of paper and review, we fine-tune a separate Qwen3-8B model to predict author agreement (i.e., validity), action, and response. We use a held-out subset of \ourdataset{} with 3K papers to train the evaluator model. The validity and action classifiers achieve F1 scores of 0.753 and 0.740 respectively on a test set of 1K feedback, with full metrics reported in Appendix~\ref{app:author-response-pred-training-results}.

\paragraph{Quality Dimension Scoring}
Before conducting author response prediction, we apply a quality filter based on LLM-as-a-judge scoring across four quality dimensions: accuracy, paper-specific grounding, constructive tone, and prioritization \citep{steiss2024comparing}—the same dimensions used in feedback corruption (\S\ref{sec:goodpoint-iclr}). While author response prediction captures bottom-up signals for feedback validity, it is trained on human-only data and may thus fail out-of-distribution on model-specific error patterns. Quality dimension scoring addresses this gap by filtering feedback before author response prediction. We derive score thresholds from average LLM-judge scores over 2K human-reviewer feedback examples, balanced across valid and invalid instances. Using GPT-5 mini (\texttt{gpt-5-mini-2025-08-07}), we score each feedback on a 5-point Likert scale given the paper content and feedback text. Prompts and implementation details are provided in Appendix~\ref{app:quality-dimension-scoring-setup}.

\subsection{Human Consensus-Based Feedback Evaluation}

We evaluate LLMs' ability to generate important feedback by using \textit{human consensus feedback} as a proxy for important feedback.
We define human consensus feedback as the set of feedback that is raised by multiple reviewers and is also \textit{successful} (i.e., valid and actionable).
We then assess whether each LLM-generated feedback matches human consensus feedback to evaluate the quality of the feedback set.
The definition and example of feedback match appear in Appendix~\ref{app:consensus-eval-example}.

\paragraph{Metric Definitions}
We use precision, recall, and F1 scores as performance measures.
Precision measures the fraction of LLM-generated feedback that matches human consensus feedback, thereby penalizing models that generate excessive non-essential feedback.
Recall measures the fraction of human consensus feedback that is matched by LLM-generated feedback.
F1 is the harmonic mean of precision and recall.
The formula of these metrics is provided in the Appendix~\ref{app:consensus-metrics-def}.

\paragraph{Feedback Match Assessment} 
We assess feedback matches in two scenarios: for human--human pairs to construct human consensus feedback, and for human--LLM pairs to evaluate LLM feedback against the human consensus feedback.
First, we filter out pairs with text embedding similarities below a specific threshold (0.55 for human--human and 0.45 for human--LLM). To determine these thresholds, we manually annotate a sample of feedback pairs and set the cutoffs to exclude similarity ranges where the annotated match rate falls below 0.1 (see Appendix~\ref{app:feedback-pair-filtering} for details).
For the remaining pairs, we use GPT-5.2 to automatically classify whether each pair is a match. Compared to human annotations on the sampled pairs, GPT-5.2 achieves F1 scores of 0.867 and 0.906 for human--human and human--LLM pairs, respectively. More details on the automated feedback matching are provided in the Appendix~\ref{app:fb-assessment-performance}.

\subsection{Human Evaluation}
For human evaluation, we conduct a user study with $N=13$ authors evaluating feedback generated on their own papers. The authors were primarily PhD students with 2-6 years of research experience. The papers included initial manuscripts submitted to ACL, EMNLP, ICLR, and ICML. The authors are asked to perform an item-level assessment in four dimensions: validity, action they will take given the feedback, and specificity and helpfulness on Likert scale. In addition to our targeted quality dimensions, validity and action, specificity and helpfulness are included for a more comprehensive evaluation, following previous work \citep{sadallah-etal-2025-good, liang2024can}. 
We provide further details in Appendix~\ref{app:human-eval}.\looseness=-1

\section{Experiment Setup \& Results}

\paragraph{Datasets}
Across experiments, we use a test set of 1,198 papers where the number of accepted and rejected papers are balanced. To build the test set, we combine 600 papers from ICLR 2020-2025 and 598 from ICLR 2026. We report the results on both the aggregated and separated test sets to address potential data contamination. In human consensus-based feedback evaluation, we use 943 papers from the test set, excluding the papers without any human consensus feedback. These papers have 5.97 consensus feedback items per paper, on average. For human evaluation, we use author-provided manuscripts. \looseness=-1

\paragraph{Baselines}
We compare our models with four baseline models across closed- and open-models of different families and sizes on quality feedback generation. For open model comparison, we use Llama 3.1-8b-Instruct and Qwen3-8B. For closed models, we compare to GPT-5.2 and Gemini-3-flash. In human evaluation, we compare one proprietary, Gemini-3-flash, and the strongest size-comparable models, Qwen3-8B, to \ours{}-DPO.
We provide models with the manuscript converted to markdown and use identical prompts across the models to generate feedback on the paper. We parse each model-generated review into single self-contained feedback units for comparison. The exact prompts and settings for generation are detailed in Appendix~\ref{app:baseline-generation}.

\paragraph{Subsampling for Length Normalization}
To mitigate potential biases arising from the number of feedback units generated by different LLMs, we randomly subsample 5 feedback units from each model's total output for a given paper before evaluating their quality. In automatic evaluations, we report the mean and 95\% confidence intervals calculated over $B=1,000$ bootstrap iterations.

\subsection{Author Response Prediction}
\label{sec:author-response-prediction-result}
Table~\ref{tab:result-author-response-prediction} reports success rates under combined, author response prediction only, and quality score filtering only metrics. GPT-5.2 achieves the highest combined success rate ($45.8\%\pm1.0\%$), followed by Gemini-3-flash ($37.9\%\pm0.9\%$). \ours{}-DPO and \ours{}-SFT outperform Qwen3-8B ($8.0\%\pm0.6\%$) by 6.7\% and 1.2\% on the combined success rate and author action only pass rate, and by 6.8\% and 1.6\% on validity only pass rate, respectively. Among size-comparable open-source models, Qwen3-8B substantially outperform Llama3.1-8B-Instruct ($1.8\pm0.3\%$). Together, these results indicate that \ours{} training meaningfully improves feedback quality.
\begin{table}[t]
\begin{center}
\small
\begin{tabular}{l cc cc cc}
\toprule
& \multicolumn{2}{c}{\bf Combined Success}
& \multicolumn{2}{c}{\bf Validity Only}
& \multicolumn{2}{c}{\bf Author Action Only} \\
\cmidrule(lr){2-3} \cmidrule(lr){4-5} \cmidrule(lr){6-7}
\bf Model & \bf Rate (\%) & \bf CI ($\pm$\%)
          & \bf Rate (\%) & \bf CI ($\pm$\%)
          & \bf Rate (\%) & \bf CI ($\pm$\%) \\
\midrule
Gemini-3-flash        & 37.9 & 0.9 & 39.4 & 0.9 & 37.9 & 0.9 \\
GPT-5.2               & \textbf{45.8} & 1.0 & \textbf{46.3} & 1.0 & 45.8 & 1.0 \\
Llama3.1-8b-Instruct  & 1.8 & 0.3 & 1.8 & 0.3 & 1.8 & 0.3 \\
Qwen3-8b (Base)       & 8.0 & 0.6 & 8.1 & 0.6 & 8.0 & 0.6 \\
\midrule
\ours{}-DPO & 14.7 (+6.7) & 0.5 & 14.9 (+6.8) & 0.5 & 14.7 (+6.7) & 0.5 \\
\ours{}-SFT & 9.2 (+1.2) & 0.5 & 9.7 (+1.6) & 0.5 & 9.2 (+1.2) & 0.5 \\
\bottomrule
\end{tabular}
\end{center}
\caption{Pass rates ($B=1000$, 5 feedback/paper) across three scoring configurations:
\textit{Combined Success} uses both validity and author action signals;\textit{Validity Only} and \textit{Author Action Only} use each signal independently.
CI half-widths are from 95\% confidence intervals; $\Delta$ for \ours{} models is relative to Qwen3-8b (Base).}
\label{tab:result-author-response-prediction}
\end{table}

\begin{table}[t]
\begin{center}
\small 
\begin{tabular}{l c c c} 
\toprule
\bf Model & \bf Precision & \bf Recall & \bf F1 \\
\midrule
Gemini-3-flash & 0.128 & \textbf{0.169} & \textbf{0.131} \\
GPT-5.2 & 0.130 & 0.165 & 0.130 \\
Llama3.1-8b-Instruct & 0.047 & 0.053 & 0.044 \\
Qwen3-8b (Base) & 0.069 & 0.084 & 0.068 \\
\midrule
\ours{}-DPO & 0.093 \scriptsize{(+0.024)} & 0.107 \scriptsize{(+0.023)} & 0.087 \scriptsize{(+0.019)} \\
\ours{}-SFT & \textbf{0.138} \scriptsize{(+0.069)} & 0.112 \scriptsize{(+0.028)} & 0.108 \scriptsize{(+0.040)} \\
\bottomrule
\end{tabular}
\end{center}
\caption{Model performances in matching human consensus feedback ($B=1000$, 5 feedback/paper).
CI half-widths computed from 95\% confidence intervals for all values are less than 0.001. Among the size-comparable models (7B/8B), \ours{}-SFT achieves the best performance, followed by \ours{}-DPO, across all metrics. Notably, \ours{}-SFT achieves higher precision than the larger proprietary models.}
\label{tab:result-consensus-fb-eval}
\end{table}

\subsection{Alignment with Human Consensus Feedback}
\label{sec:consensus-eval-result}
Table \ref{tab:result-consensus-fb-eval} presents the performance of \ours{} models and baselines in matching human consensus feedback. Among comparably sized open-weight models, \ours{}-SFT achieves the best performance across all metrics, substantially narrowing the gap to much larger proprietary models. Compared to its base model, Qwen3-8B—also the strongest open-weight baseline—\ours{}-SFT yields a 58.8\% improvement in F1 (0.108 vs.\ 0.068) and doubles precision (0.138 vs.\ 0.069).

Notably, \ours{}-SFT achieves higher precision than the larger proprietary models, Gemini-3-flash (0.128) and GPT-5.2 (0.130). This suggests that \ours{} produces more selective and high-fidelity critiques with fewer low-quality comments, despite its substantially smaller scale. \ours{}-DPO follows closely, likewise outperforming all similarly sized baselines across every metric. These findings remain consistent across the two test partitions, including the temporally held-out ICLR 2026 set (Appendix~\ref{app:cons-eval-results-partition}). Finally, bootstrap resampling (B=1,000) confirms the statistical robustness of these results, with confidence-interval half-widths consistently below 0.001 for all metrics.

\subsection{Human Evaluation}
Table~\ref{tab:human-eval-results} summarizes the human evaluation results. Gemini-3-flash achieves the strongest performance in all four dimensions, and \ours{}-DPO ranks second overall, outperforming Qwen3-8B in all dimensions: validity (58.1\% vs.\ 41.5\%), actionability (40.3\% vs.\ 32.3\%), specificity (3.50 vs.\ 2.89), and helpfulness (2.77 vs.\ 2.25). These results indicate that \ours{}-DPO meaningfully improves author-perceived feedback quality, narrowing the gap to the strongest proprietary baseline. These findings are also consistent with our automatic evaluation, suggesting that the improvements are not limited to automatic metrics but also translate into practical usefulness for real scientific paper authors.

\begin{table}[t]
\centering
\small
\begin{tabular}{lccccc}
\toprule
\textbf{Model} & \textbf{Validity} & \textbf{Actionability} & \textbf{Specificity} & \textbf{Helpfulness} & \textbf{$n$} \\
& \textbf{Rate} & \textbf{Rate} & \textbf{Mean (SD)} & \textbf{Mean (SD)} & \\
\midrule
Gemini-3-flash & $72.3\%^{\dagger\ddagger}$ & $56.9\%^{\ddagger}$ & $4.42\ (0.86)^{\dagger\ddagger}$ & $3.40\ (1.40)^{\dagger\ddagger}$ & 65 \\
\ours{}-DPO    & $58.1\%$                  & $40.3\%$            & $3.50\ (1.16)^{\ddagger}$       & $2.77\ (1.29)^{\ddagger}$       & 62 \\
Qwen3-8B       & $41.5\%$                  & $32.3\%$            & $2.89\ (1.28)$                  & $2.25\ (1.24)$                  & 65 \\
\bottomrule
\end{tabular}
\caption{Human evaluation at the feedback level ($n$ per model shown). Superscripts denote significant pairwise differences: ${\dagger}$~vs.\ \ours{}-DPO, ${\ddagger}$~vs.\ Qwen3-8B. Ordinal metrics (Specificity, Helpfulness) use Kruskal--Wallis with Mann--Whitney~U pairwise tests; binary metrics (Validity, Actionability) use $\chi^2$ with Fisher's exact pairwise tests. }
\label{tab:human-eval-results}
\end{table}




\section{Analysis \& Discussion}
To better understand the role of LLMs in feedback generation, we analyze both their limitations relative to human and their potential to complement human feedback. Specifically, we first present our analysis of LLM failure modes through author response distribution in human evaluation and quality scoring. We then turn to a complementary question: whether LLMs can provide novel feedback beyond what human reviewers typically offer.

\subsection{LLM Generated Feedback Failure Modes}
One of the main immediately noticeable differences in human and LLM generated feedback includes their length. The number of feedback generated vary where human reviewer generated 5.66 feedback units across \ourdataset{} while LLM generated feedback numbers varied with highest number generated by Qwen-3-8B (20.85), followed by GPT-5.2 (20.72), Llama3.1-8B-Instruct (11.96), and Gemini-3-flash (10.79), suggesting possible verbosity of model-generated feedback. On the other hand, \ours{}-SFT and \ours{}-DPO generated 4.15 and 5.71 average feedback units respectively, more closely following human distribution but could indicate lower coverage. 


\begin{figure}[b]
    \centering
    \includegraphics[width=0.7\textwidth]{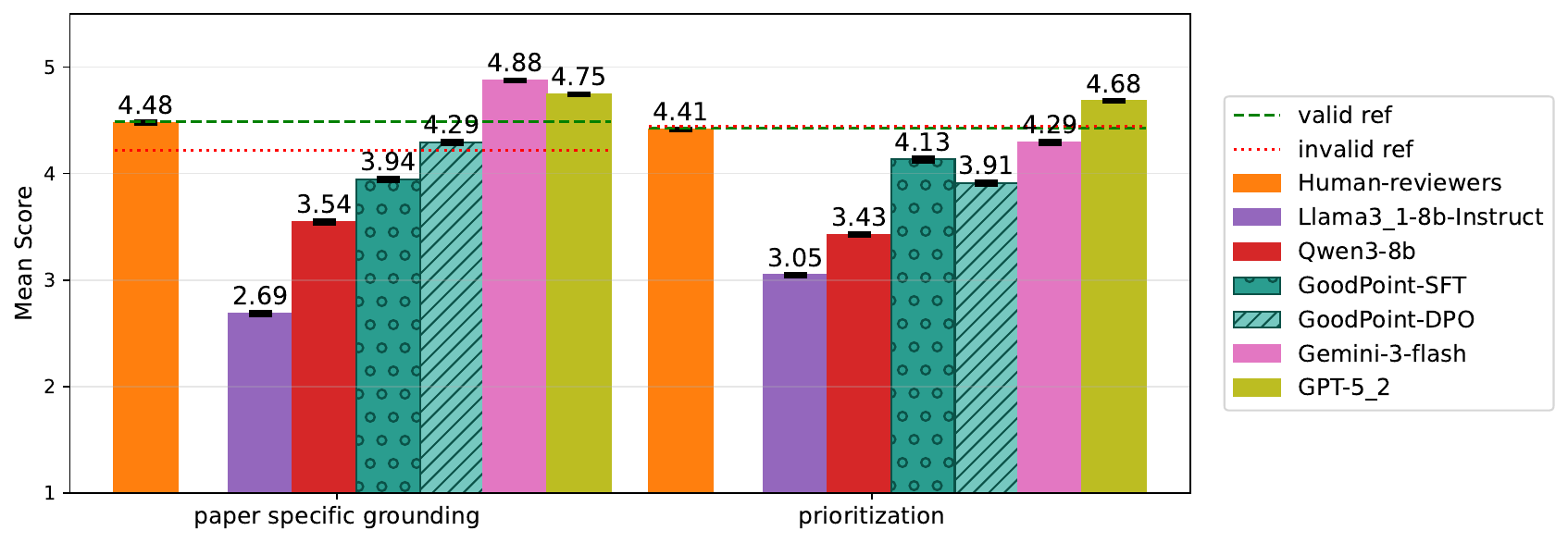}

    \vspace{\baselineskip}
    \caption{Mean scores for paper-specific grounding and prioritization across models with SEM. The dotted lines show reference scores of 1000 valid and invalid feedback items each.}
    \label{fig:quality-scores-bar}
\end{figure}

\textbf{LLMs generate bland and general feedback without training.} One noticeable failures of LLM-generated feedback is lack of specificity, especially for smaller (7-8B) open-sourced models. This is evident in both quality scoring in \S\ref{sec:evaluation} and human evaluation responses. As shown in Figure~\ref{fig:quality-scores-bar}, both Qwen3-8B and Llama3.1-8B-Instruct show mean paper specific grounding and prioritization scores considerably below human scores and reference score obtained on 1000 invalid reviewer feedback items. This is corroborated by human annotation results both quantitatively (Table~\ref{tab:human-eval-results}) and qualitatively by participant provided details on feedback generated by Qwen3-8B (``\textit{not specifc[sic] enough to make revision or judge whether it is valid or not}'', P7). Through training, \ours{} models achieve higher scores for specificity (+0.75; \ours{}-DPO) and prioritization (+0.7; \ours{}-SFT). 

\begin{figure}[t]
    \centering
    \includegraphics[width=\textwidth]{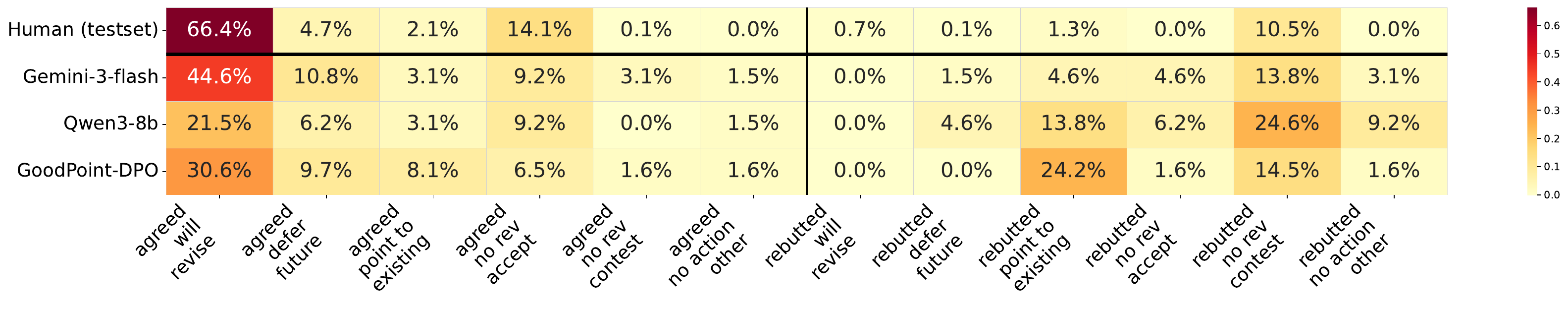}
    \caption{Author action and validity distributions from human evaluation. The top row shows results from human reviewer/author baselines on the test set. `No rev' = no revision.}
    \label{fig:author-action-heatmap-human-eval}
\end{figure}

\textbf{LLMs miss existing content or hallucinate more than human reviewers.}
As shown in Figure~\ref{fig:author-action-heatmap-human-eval}, the author response distribution shows that \ours{}-DPO outperforms base model in both validity (agreed) and actionability (will revise, defer to future work). Moreover, our results highlight some key failure modes of LLMs in missing already addressed details. For human reviewers, point to existing content action only occurs 2.1\% when feedback is valid and 1.3\% when invalid. However, in LLMs, this rate increases to 3.1-8.1\% when valid and 4.6-24.2\% when invalid. This could indicate that models struggle to handle and provide feedback for long-context input, highlighting a key improvement area. Hallucination was another key issue noted by participants during human evaluation (\textit{``This would read as LLM-hallucinated feedback.''}, P5 on Gemini-3-flash generated feedback). 

\subsection{Human Alignment vs. Novel Feedback Generation}
\label{sec:human-alignment-novel-fb}
While alignment with reviewer consensus is important, LLMs can also offer complementary feedback beyond what human reviewers typically mention. To study this, we decompose the success rate improvement of \ours{} over the base model into two components: \textit{aligned feedback} (\textit{Matched \& Successful}) and \textit{novel feedback} (\textit{Non-matched \& Successful}), based on our automatic evaluations (\S\ref{sec:evaluation}). We further categorize novel feedback into 11 aspects following \cite{liang2024can} using LLM-based classification.\looseness=-1

\textbf{DPO drives novelty, while SFT drives alignment.} Both \ours{}-DPO and \ours{}-SFT show larger relative gains in aligned feedback rate (i.e., the proportion of aligned feedback among all generated feedback) than in novel feedback rate (i.e., the proportion of novel feedback among all generated feedback) (\ours{}-DPO: 0.93 vs.\ 0.88; \ours{}-SFT: 0.64 vs.\ 0.24). However, \ours{}-DPO is more balanced, likely because SFT directly imitates successful human feedback, whereas DPO optimizes preferences over broader quality dimensions. This matches our automatic evaluations: 
\ours{}-SFT outperforms \ours{}-DPO on consensus match (\S\ref{sec:consensus-eval-result}) but not on success rate (\S~\ref{sec:author-response-prediction-result}).
Furthermore, \ours{}-SFT’s gains in novel feedback are concentrated in \textit{Add experiments on More Datasets}, whereas \ours{}-DPO shows gains across broader aspects (e.g., \textit{Add Ablations Experiments}, \textit{Algorithm Efficiency}, and \textit{Theoretical Soundness}), suggesting that DPO promotes a more generalized capacity for novel feedback generation (see Appendix~\ref{app:novel-feedback} for details). \looseness=-1

\section{Conclusion}
\label{sec:discussion-conclusion}
We present \ours{}, a framework for generating constructive, author-centric feedback by leveraging signals from real-world author responses. By defining feedback quality along the axes of validity and actionability, we developed both a large-scale dataset and a training strategy that aligns LLM outputs with feedback that authors acknowledge and act upon.
Experiment results on both automatic metrics and human evaluation show that our approach delivers high-quality, precise feedback, even enabling small models to rival much larger ones.
Our findings highlight the importance of grounding LLMs in human-centric signals.\looseness=-1

\section*{Limitations}
Our approach uses author responses as a supervision signal, but these are an imperfect proxy for feedback quality, as replies can be shaped by time constraints, community norms, and rebuttal incentives; while we mitigate this with DPO and quality filtering, future work could explore expert audits, finer-grained actionability labels, or revision-based outcome signals. Additionally, our focus on ICLR limits generalizability, as feedback norms and criteria may differ across venues and disciplines, warranting future validation across additional fields.
\bibliography{colm2026_conference,custom}
\bibliographystyle{colm2026_conference}

\appendix
\section{Extended Related Works}
\label{app:extended-related-works}
\paragraph{LLMs in Scientific Research}
LLMs are increasingly used in scientific workflows, including literature synthesis, hypothesis generation, and manuscript writing \citep{chen2025ai4research, huang2025deep, si2025ideation}. However, the growing push for fully autonomous research systems \citep{ifargan2025autonomous, schmidgall2025agent} may distort scientific judgment and create harmful incentive shifts in the research ecosystem \citep{baumannstop, Wilder_2025}. We advocate for an augmentation approach, where LLMs collaborate with researchers, enhancing their work without replacing their critical judgment.

\paragraph{Automated Peer Review}
Automated peer review has gained interest as a solution to reviewer burden and inconsistency \citep{zhou2024llm, liang2024can, idahl-ahmadi-2025-openreviewer}. However, recent studies reveal limitations of llm-generated peer reviews, including their struggle with score prediction and providing human-level critical feedback \citep{zhou2024llm} and being overly positive \citep{idahl-ahmadi-2025-openreviewer}.  These works typically evaluate quality using reviewer-centric metrics like aspect coverage and score prediction. Existing peer review generation methods include fine-tuning language models on expert reviews \citep{idahl-ahmadi-2025-openreviewer}, standardizing multiple reviews for supervision \citep{yu-etal-2024-automated}, and using structured analysis and literature retrieval for evidence-based reviewing \citep{zhu2025deepreview}.
Our focus is on generating constructive feedback to improve papers, rather than evaluating them for acceptance. Unlike prior systems that rely on fixed criteria or full human reviews, \ours{} uses authors' feedback acknowledgments as a training signal for author-centric feedback generation.

\begin{table}[h]
\centering
\small
\begin{tabular}{lcccc}
\toprule
\textbf{Label} & \textbf{Observed Agr.} & \textbf{PABAK} & \textbf{Cohen's $\kappa$} & \textbf{$n$} \\
\midrule
Validity        & 0.874 & 0.748 & 0.000 & 111 \\
Author Action   & 0.919 & 0.838 & 0.271 & 111 \\
\bottomrule
\end{tabular}
\caption{Inter-annotator agreement for validity and author action labels ($n=111$ units).
PABAK is preferred over Cohen's $\kappa$ due to high label skew
\citep{byrt1993bias}.}
\label{tab:iaa}
\end{table}

\section{Parsing Setup \& Performance}
\label{app:parsing-setup}

To parse human reviews into self-contained feedback units with corresponding author responses with validity and author action, we use GPT4.1 (\texttt{gpt-4.1-2025-04-14}). We set temperature to 0.7 and max generation toke length to 10240. We also annotate feedback dimensions \citep{hattie2007power} and key aspects as defined in \citeauthor{liang2024can}. Two members of the research team annotated validity and author action. The overall agreements between annotators are shown in Table~\ref{tab:iaa}. The user prompt below is summarized due to length. 

\begin{tcolorbox}[
    colback=gray!5,
    colframe=gray!40,
    fontupper=\ttfamily\small,
    title={\textbf{System Prompt}},
    breakable
]
You are a helpful assistant, an expert at analyzing academic peer review responses and extracting feedback.
\end{tcolorbox}

\begin{tcolorbox}[
    colback=gray!5,
    colframe=gray!40,
    fontupper=\ttfamily\small,
    title={\textbf{User Prompt}},
    breakable
]
You will receive a conversation between *reviewers* and *authors* for an academic paper.
Your task is to **parse each reviewer feedback item** into a **self-contained unit** and
evaluate the **corresponding author response**. Ignore purely positive feedback with no
actionable critique, question, or suggestion.

For **each feedback unit**, extract and annotate the following fields:

\smallskip
\textbf{1. feedback\_text} --- Rewrite the reviewer feedback as a clear, standalone unit,
preserving the original text as closely as possible.

\smallskip
\textbf{2. author\_response\_text} --- Rewrite the author response as a clear, standalone
unit, preserving tone and first-person pronouns.

\smallskip
\textbf{3. validity} --- Did the authors agree the feedback is valid?
\texttt{agreed\_by\_authors} | \texttt{rebutted\_by\_authors} | \texttt{unclear}

\smallskip
\textbf{4. author\_action} --- What did the authors commit to do?
\texttt{will\_revise} | \texttt{defer\_future\_work} | \texttt{point\_to\_existing\_content}
| \texttt{no\_revision\_accept} | \texttt{no\_revision\_contest} | \texttt{no\_action\_other}
| \texttt{unclear\_or\_no\_response}

\smallskip
\textbf{5. dimensions} --- Break down the feedback into three dimensions if present:
\textit{Feed Up} (goal), \textit{Feed Back} (gap), \textit{Feed Forward} (next steps).

\smallskip
\textbf{6. aspects} --- Identify emphasized aspects (e.g., Clarity and Presentation,
Reproducibility, Novelty, Theoretical Soundness, \textit{etc.})

\smallskip
Return output strictly as JSON following the prescribed schema. Full decision rules,
label definitions, and output schema are omitted here for brevity.

\smallskip
\textbf{Input:} \{conversation\_text\}
\end{tcolorbox}

\section{Feedback Corruption \& Verification}
\label{app:feedback-corruption}

\paragraph{Feedback corruption} To teach models high quality feedback through meaningful contrastive pairs, we generate feedback corrupted set. We use GPT-5 mini (\texttt{gpt-5-mini-2025-08-07}, medium reasoning and medium verbosity, 4096 max tokens) to generate corrupted set in each dimension. 
\begin{tcolorbox}[
    colback=gray!5,
    colframe=gray!40,
    fontupper=\ttfamily\small,
    title={\textbf{System Prompt}},
    breakable
]
The model acts as a text transformation assistant that rewrites reviewer feedback by deliberately corrupting exactly one quality dimension at a time, while preserving all others.
\end{tcolorbox}

\begin{tcolorbox}[
    colback=gray!5,
    colframe=gray!40,
    fontupper=\ttfamily\small,
    title={\textbf{User Prompt}},
    breakable
]
\textbf{Inputs:} Paper title, abstract, and original reviewer feedback.

\medskip
Given these inputs, the model produces five rewrites of the feedback, each degrading a single dimension:

\medskip
\begin{tabularx}{\linewidth}{lX}
\toprule
\textbf{Corruption} & \textbf{Description} \\
\midrule
Generic & Remove all paper-specific details so the feedback reads as boilerplate applicable to any paper in the field. \\
Vague & Retain paper-specific references but strip concrete questions, examples, and actionable guidance. \\
Inaccurate & Introduce plausible but factually wrong claims about the paper's content, methods, or results. \\
Non-essential & Shift focus from core fixable issues to peripheral, out-of-scope, or stylistic concerns. \\
Unsupportive & Replace constructive, hedged language with blunt, dismissive, or commanding phrasing. \\
\bottomrule
\end{tabularx}

\medskip
Output is a JSON object with five keys (\texttt{generic}, \texttt{vague}, \texttt{inaccurate}, \texttt{nonessential}, \texttt{unsupportive}), each containing the full rewritten feedback as a string.
\end{tcolorbox}
\begin{table}[h]
\centering
\small
\caption{Corruption quality verification results by dimension. Accuracy measures whether the verifier correctly identifies the targeted dimension. Degradation and collateral preservation scores are on a 1--3 scale.}
\label{tab:corruption-verification}
\begin{tabular}{lccc}
\toprule
\textbf{Dimension} & \textbf{Accuracy} & \textbf{Target Degradation} & \textbf{Collateral Preservation} \\
\midrule
Generic       & 0.625 & 1.83 & 2.92 \\
Vague         & 0.945 & 1.86 & 2.98 \\
Inaccurate    & 0.992 & 2.98 & 2.48 \\
Non-essential & 0.969 & 3.00 & 2.56 \\
Unsupportive  & 0.982 & 2.68 & 2.88 \\
\bottomrule
\end{tabular}
\end{table}
\paragraph{Corruption verification} In addition to feedback corruption, to ensure quality of corrupted samples, we used LLM-as-a-judge for verification. We remove any samples with erroneous corrupted dimension prediction and corruption and collateral preservation score below 2. Verification results, prior to filtering, are shown in Table~\ref{tab:corruption-verification}. 

\begin{tcolorbox}[
    colback=gray!5,
    colframe=gray!40,
    fontupper=\ttfamily\small,
    title={\textbf{System Prompt}},
    breakable
]
The model acts as a judge of AI-generated text transformations in academic peer review, identifying which quality dimension has been corrupted and evaluating the quality of the corruption.
\end{tcolorbox}

\begin{tcolorbox}[
    colback=gray!5,
    colframe=gray!40,
    fontupper=\ttfamily\small,
    title={\textbf{User Prompt}},
    breakable
]
\textbf{Inputs:} Paper title, abstract, original reviewer feedback, and a randomized list of rewritten feedback variants (corruption dimensions unknown).

\medskip
For each rewrite, the model predicts the targeted corruption dimension and scores it on two criteria:

\medskip
\begin{tabularx}{\linewidth}{lX}
\toprule
\textbf{Score} & \textbf{Description} \\
\midrule
Target Degradation (1--3) & How clearly the intended dimension is degraded (3 = clearly degraded, 1 = barely degraded). \\
Collateral Preservation (1--3) & How well all other dimensions are preserved (3 = fully preserved, 1 = multiple dimensions affected). \\
\bottomrule
\end{tabularx}

\medskip
Output is a JSON object with a \texttt{results} array ordered by rewrite index, each entry containing: \texttt{rewrite\_index}, \texttt{predicted\_dimension}, \texttt{target\_degradation\_score}, \texttt{collateral\_preservation\_score}, and \texttt{reasoning}.
\end{tcolorbox}

\section{Training Setup}
\label{app:training-setup}
\ours{}-SFT is fine-tuned from \texttt{Qwen3-8B} and \ours{}-DPO is subsequently
initialized from \ours{}-SFT, both trained using OpenRLHF \citep{hu2024openrlhf} with the following
configurations:

\begin{table}[h]
\centering
\small
\begin{tabular}{lll}
\toprule
\textbf{Hyperparameter} & \textbf{SFT} & \textbf{DPO} \\
\midrule
Base model              & Qwen3-8B        & \ours{}-SFT \\
Max sequence length     & 30{,}000        & 30{,}000 \\
Train batch size        & 128             & 128 \\
Micro batch size        & 8               & 4 \\
Learning rate           & $5 \times 10^{-6}$ & $5 \times 10^{-6}$ \\
Epochs                  & 1               & 1 \\
Precision               & BF16            & BF16 \\
\midrule
$\beta$ (KL penalty)    & ---             & 0.1 \\
NLL loss coefficient    & ---             & 0.2 \\
\bottomrule
\end{tabular}
\caption{SFT and DPO training hyperparameters.}
\label{tab:training-config}
\end{table}

For GPU optimization, we use ZeRO stage 3 with flash-attention-2 with ring attention size 8 with head stride 2. We also early stop DPO at step 50 to prevent overfitting. 

For \ours{}-DPO, we also use cosine similarities of embeddings using OpenAI's \texttt{text-embedding-3-small} to deduplicate multiple overlapping reviewer feedback. We randomly select among the matched instances if the similarity is greater than 0.5.

\section{Author Response Prediction Training \& Results}
\label{app:author-response-pred-training-results}

\begin{table}[h]
\centering
\small
\caption{Author response prediction results on a held-out test set of 1,000 samples (864 valid, 136 invalid).}
\label{tab:author-response-prediction-performance}
\begin{tabular}{lcccc}
\toprule
\textbf{Task} & \textbf{Precision} & \textbf{Recall} & \textbf{F1} & \textbf{Accuracy} \\
\midrule
Validity & 0.815 & 0.717 & 0.754 & 0.717 \\
Action   & 0.779 & 0.706 & 0.741 & 0.650 \\
\bottomrule
\end{tabular}
\end{table}

\subsection{Author Response Predictor Training}
Author response predictor is trained on 3K papers and their corresponding feedback items, author validity, action, and response. We balance valid (agreed) and invalid (rebutted) feedback items to address label imbalance in the dataset. Due to the lack of invalid feedback samples, we supplement our 3K paper set with filtered out unused invalid feedback samples from SFT data. We also used OpenRLHF \citep{hu2024openrlhf} for training with the following hyperparameters: a maximum sequence length of 30,000 tokens, a train batch size of 32 (micro batch size of 1), and a learning rate of $5 \times 10^{-6}$ for 1 epoch in BF16 precision. Distributed training uses ZeRO stage 3 with FlashAttention-2, ring attention of size 8 (head stride 2), sample packing, gradient checkpointing, and Adam offload. Table~\ref{tab:author-response-prediction-performance} shows full performance metric over 1000 test samples with 864 valid and 136 invalid samples. 

\subsection{Quality Dimension Scoring Setup}
\label{app:quality-dimension-scoring-setup}
We obtain quality scores for feedback units using GPT-5 mini (\texttt{gpt-5-mini-2025-08-07}). We set verbosity and reasoning effort to medium and max new token length to be 4096. We use the following prompts (truncated and summarized for space):

\begin{tcolorbox}[
    colback=gray!5,
    colframe=gray!40,
    fontupper=\ttfamily\small,
    title={\textbf{System Prompt}},
    breakable
]
The model acts as an expert meta-reviewer evaluating peer-review feedback impartially and evidence-based, citing specific phrases to justify every score.
\end{tcolorbox}

\begin{tcolorbox}[
    colback=gray!5,
    colframe=gray!40,
    fontupper=\ttfamily\small,
    title={\textbf{User Prompt}},
    breakable
]
Given a paper excerpt, venue, and reviewer feedback, the model scores feedback across five dimensions (1--5 each):

\medskip
\begin{tabularx}{\linewidth}{lX}
\toprule
\textbf{Dimension} & \textbf{Description} \\
\midrule
Accuracy & Are claims about the paper's content, methods, and results factually correct? \\
Prioritisation & Does feedback focus on high-impact scientific issues (validity, novelty, reproducibility) over cosmetic ones? \\
Constructive Tone & Is feedback collegial, respectful, and improvement-oriented? \\
Paper-Specific Grounding & Is feedback anchored to named components, results, and claims of this paper vs.\ generic boilerplate? \\
Actionability & Are revision directions specific, feasible, and tied to concrete sections or analyses? \\
\bottomrule
\end{tabularx}

\medskip
Output is a structured JSON object containing per-dimension scores, justifications, and supporting evidence, plus an overall summary with average score, strengths, weaknesses, and holistic judgment.
\end{tcolorbox}

\begin{figure}[!htb]
    \centering
    \begin{minipage}[t]{0.7\textwidth}
        \vspace*{0mm}
        \includegraphics[width=\textwidth]{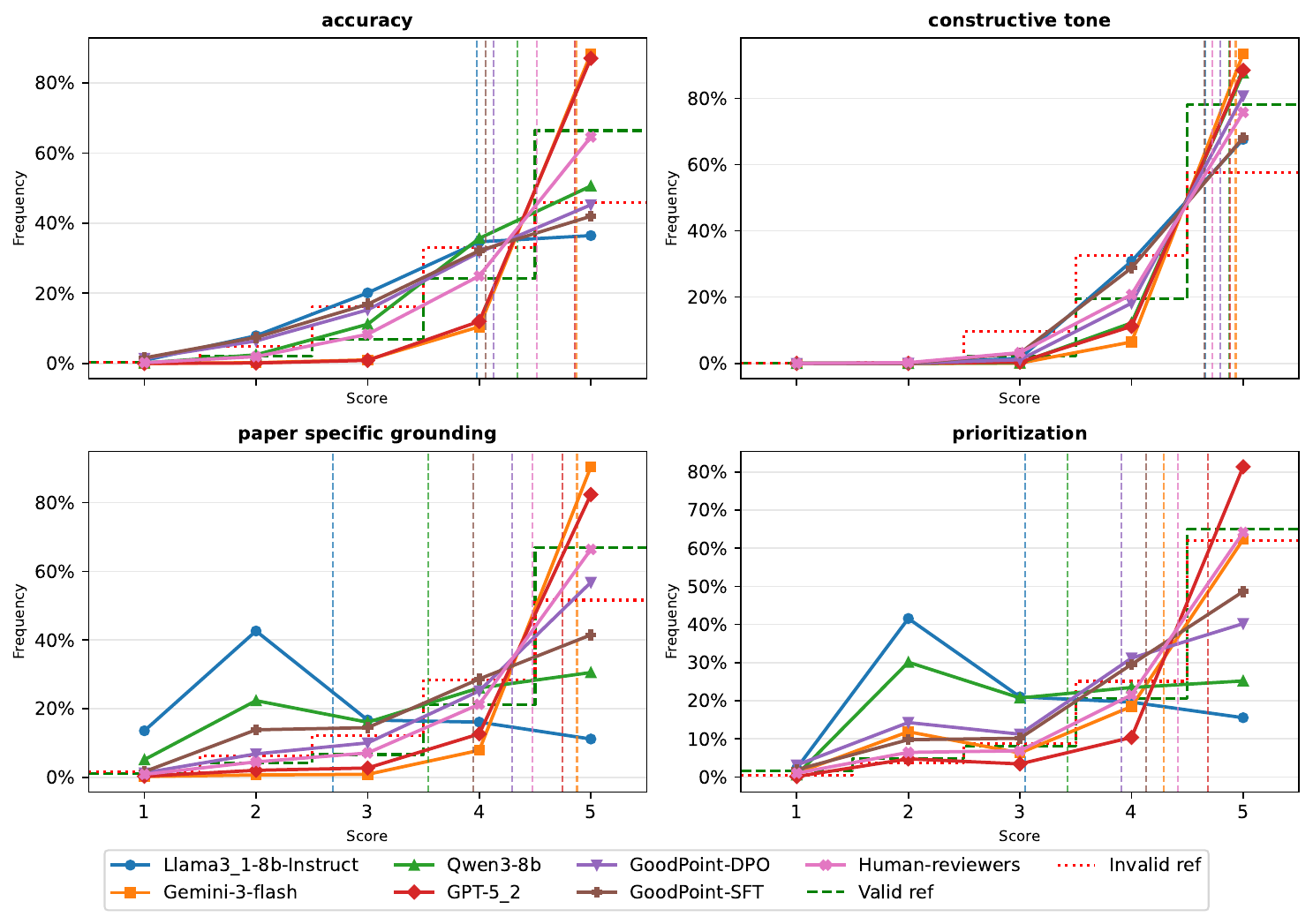}
    \end{minipage}
    \hfill
    \begin{minipage}[t]{0.28\textwidth}
        \vspace*{0mm}
        \caption{\raggedright LLM-judged feedback quality score distributions for baseline models, \ours{} models, and human reviewers on test set. Valid ref and Invalid ref indicate scores on 1000 valid and invalid feedback units respectively. All dimensions are scored on a Likert scale from 1 to 5. Dotted vertical lines indicate mean scores.}
        \label{fig:quality-scores}
    \end{minipage}
\end{figure}

Figure~\ref{fig:quality-scores} show distribution of quality scores across models and humans. We see that LLMs specifically have noticeably lower scores in paper-specific grounding and prioritization. Furthermore, accuracy is also relatively lower for LLM-generated feedback. Therefore, to address these LLM-specific failure modes not observed in human data, we utilize LLM-as-a-judge quality scoring to filter based on human average scores to filter acceptable quality feedback. Across the five evaluation dimensions, the human reviewer feedback scores averaged 4.37 for accuracy, 4.44 for prioritisation, 4.62 for constructive tone, and 4.36 for paper-specific grounding, with actionability receiving the lowest score of 3.49 (all out of 5).

\begin{figure}[!htb]
    \centering
    \includegraphics[width=\textwidth]{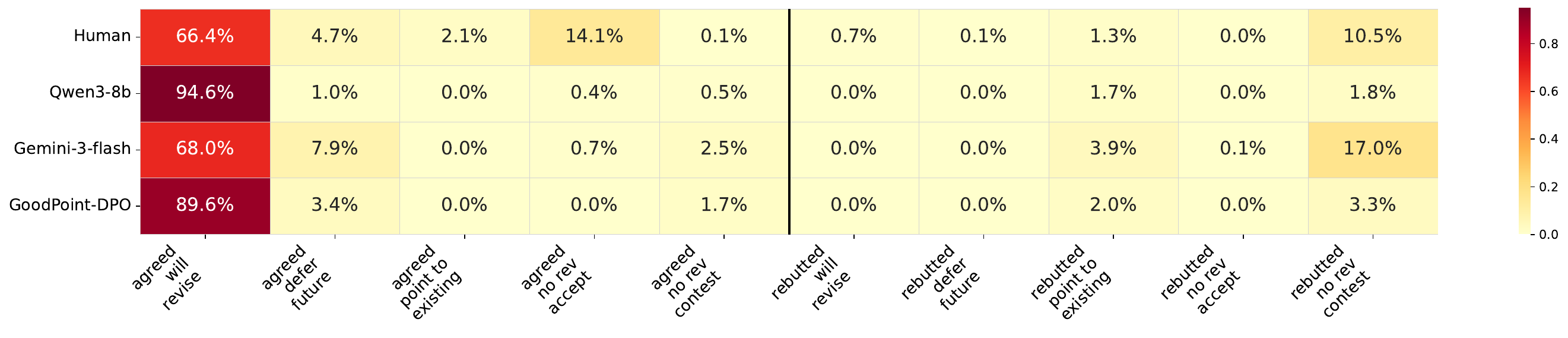}
    \caption{Human reviewer feedback author response distribution from entire \ourdataset{}. The heatmap shows percentage taken over all feedback. We see the highest concentration of author agreed with committed revisions.}
    \label{fig:author-action-heatmap-testset}
\end{figure}

\subsection{Author Response Prediction Results}
Figure~\ref{fig:author-action-heatmap-testset} shows validity and author action distribution over test set across different models and human reviewers. We observe that Gemini-3-flash has most similar distribution as the human reviewers. We also observe that Qwen3-8b and \ours{}-DPO show relatively high validity and actionability possibly due to spurious acceptance of generic feedback units. 

\section{Human Consensus-based Feedback Evaluation}

\subsection{Definition and Example of Feedback Match}
\label{app:consensus-eval-example}
In human consensus-based feedback evaluation, we consider a pair of feedback as a match if the two items include at least one concrete point that shares a targeted part of the paper, a deficiency pointed out, the quality dimension addressed, and the action requested.

Figure~\ref{fig:consensus-eval-example} shows example feedback matches for feedback sets provided by human reviewers and LLM.
Human consensus feedback is constructed by labeling matches between human feedback.
Then matches between each human consensus feedback and LLM-generated feedback are labeled to evaluate LLM's ability to generate essential human feedback.
\newlength{\tikzunit}
\setlength{\tikzunit}{0.07813\linewidth}

\begin{figure}[t]
\centering
\begin{tikzpicture}[
    x=\tikzunit,
    y=1cm,
    every node/.style={font=\small},
    item/.style={
        anchor=north west,
        text width=0.42\linewidth,
        align=left,
        inner sep=4pt
    },
    consensus/.style={
        anchor=north west,
        text width=0.42\linewidth,
        align=left,
        inner sep=4pt,
        font=\small\bfseries
    },
    llmitem/.style={
        anchor=north west,
        text width=0.41\linewidth,
        align=left,
        inner sep=4pt
    },
    hdr/.style={
        font=\small\bfseries,
        anchor=north west
    }
]

\def\lx{0.6}
\def\rx{7.4}
\def\sep{6.1}

\pgfmathsetmacro{\pitch}{1.4}   

\pgfmathsetmacro{\ya}{-0.75}
\pgfmathsetmacro{\yb}{\ya - \pitch}
\pgfmathsetmacro{\yc}{\yb - \pitch}
\pgfmathsetmacro{\yd}{\yc - \pitch}
\pgfmathsetmacro{\ye}{\yd - \pitch}

\node[hdr] at (\lx, 0) {Human Reviewers' Feedback};
\node[hdr] at (\rx, 0) {LLM-Generated Feedback};

\draw[gray!50, thin] (\lx, -0.55) -- (6.7, -0.55);
\draw[gray!50, thin] (\rx, -0.55) -- (12.8, -0.55);

\node[consensus] (h1) at (\lx, \ya)
    {R1: The presentation quality can be improved, as some of the notations are abused or not properly defined.};

\node[item]      (h2) at (\lx, \yb)
    {R1: Can the authors verify that the learned RL policy is stable in the experiments?};

\node[consensus] (h3) at (\lx, \yc)
    {R2: The paper's presentation is poor; the definitions of model error is missing, notations are inconsistent, \ldots};

\node[item]      (h4) at (\lx, \yd)
    {R2: The paper is missing discussion of existing literature on RL with stability or safety guarantees.};

\node[consensus] (h5) at (\lx, \ye)
    {R3: How do you evaluate the model learning error and how is the reference system learned?};

\node[llmitem] (l1) at (\rx, \ya)
    {The abstract should explicitly state the assumptions made and the precise stability guarantees proven.};

\node[llmitem] (l2) at (\rx, \yb)
    {Make a clear distinction between the real state and the reference state throughout the proof of Theorem 2.};

\node[llmitem] (l3) at (\rx, \yc)
    {Make the algorithm precise by specifying how the model error is computed and the reference model update steps.};

\node[llmitem] (l4) at (\rx, \yd)
    {The introduction should mention the key challenges in applying advanced RL methods to control systems.};

\draw[gray!70, thick, dashed]
    ($(h1.west) + (-0.05, 0.10)$) --
    ($(h1.west) + (-0.40, 0.10)$) --
    ($(h3.west) + (-0.40, 0.10)$) --
    ($(h3.west) + (-0.05, 0.10)$);

\draw[gray!70, thick, dashed]
    ($(h3.west) + (-0.05, -0.10)$) --
    ($(h3.west) + (-0.40, -0.10)$) --
    ($(h5.west) + (-0.40, 0.10)$) --
    ($(h5.west) + (-0.05, 0.10)$);

\draw[gray!70, thick, out=0, in=180]
    ($(h1.east) + (0.10, 0.10)$) to ($(l3.west) + (-0.05, 0.15)$);

\draw[gray!70, thick, out=0, in=180]
    ($(h5.east) + (0.10, 0.10)$) to ($(l3.west) + (-0.05, -0.15)$);

\end{tikzpicture}
\caption{Illustration of human consensus-based feedback evaluation with example feedback sets.
Matches between human feedback (dashed lines) are labeled to construct human consensus feedback (bold), defined as the human feedback agreed upon multiple reviewers (R1, R2, and R3).
Then each human consensus feedback is matched (curved line) by LLM-generated feedback to estimate LLM's ability to generate important human feedback.
}
\label{fig:consensus-eval-example}
\end{figure}
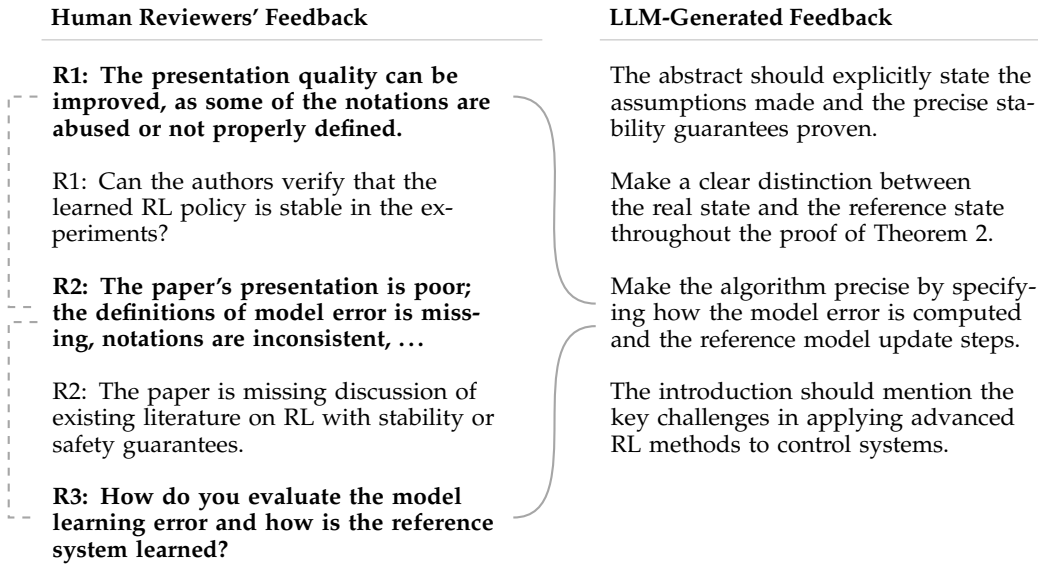

\subsection{Consensus Feedback Metrics}
\label{app:consensus-metrics-def}
\begin{table}[t]
\centering
\small
\begin{tabular}{lc}
\toprule
\textbf{Metric} & \textbf{Formula} \\
\midrule
Precision & $\dfrac{|\mathcal{M}|}{|\mathcal{F}_{\text{LLM}}|}$ \\
\addlinespace[6pt]
Recall & $\dfrac{|\mathcal{G} \cap \mathcal{C}|}{|\mathcal{G}|}$ \\
\addlinespace[6pt]
F1 & $\dfrac{2 \cdot \text{Precision} \cdot \text{Recall}}{\text{Precision} + \text{Recall}}$ \\
\bottomrule
\end{tabular}
\caption{Formula of consensus feedback metrics. $\mathcal{M}$: LLM feedback units matched by human consensus feedback; $\mathcal{F}_{\text{LLM}}$: all LLM feedback; $\mathcal{G}$: human consensus feedback; $\mathcal{C}$: human consensus feedback units matched by LLM feedback.}
\label{tab:metrics}
\end{table}
Table~\ref{tab:metrics} shows the formula of each consensus feedback metric used in the human consensus-based feedback evaluation.

\subsection{Embedding Similarity-based Feedback Pair Filtering}
\label{app:feedback-pair-filtering}
Before automatically matching human--human and human--LLM feedback pairs using GPT-5.2, we filter the pairs based on the cosine similarity of their text embeddings. We determined the thresholds for this filtering based on human annotation results from a sample of feedback pairs.
Specifically, to compare human-annotated match rates across different levels of similarity, we stratified the feedback pairs from 40 sample papers into seven cosine similarity ranges: $<$0.15, 0.15–0.25, 0.25–0.35, 0.35–0.45, 0.45–0.55, 0.55–0.65, and $>$0.65. From each stratum, we randomly sampled 20 human--human pairs and 20 human--LLM pairs, resulting in a total of 140 pairs for each category.

Each feedback pair was independently annotated by three annotators with more than three years of research experience. The final match label for each pair was determined by a majority vote. The annotation guidelines provided to them were as follows:
\begin{mdframed}[
    backgroundcolor=gray!15, 
    hidealllines=true,      
    innerleftmargin=12pt,    
    innerrightmargin=12pt,  
    innertopmargin=12pt,    
    innerbottommargin=12pt, 
    skipabove=1em,          
    skipbelow=1em           
]
\sffamily \small 

\noindent\textbf{Definition of a match between two feedback units:} \\[0.5em]
The two feedback units include at least one concrete point that shares all of the following aspects:
\begin{itemize}
    \item \textbf{Targeted part of paper}
    \begin{itemize}
        \item \textit{Detailed Guideline for Reference Resolution:} When a feedback unit references a specific element without an explicit section (e.g., ``Proposition 2''), infer the likely section based on academic conventions and context (e.g., Proposition 2 $\rightarrow$ Method/Theory). If the inferred section aligns with the section referenced in the other feedback unit, treat them as the same targeted part.
    \end{itemize}
    \item \textbf{Deficiency pointed out}
    \item \textbf{Quality dimension addressed}
    \item \textbf{Action requested}
\end{itemize}
\end{mdframed}
The annotated match rate for each stratum is presented in Table~\ref{tab:annotated-match-rate-by-stratum}. The inter-annotator agreement was moderate The inter-annotator agreement was moderate (Krippendorff’s $\alpha = 0.642$ for human--human pairs and $0.655$ for human--LLM pairs), reflecting the intrinsic ambiguity of match judgments due to differences in granularity and subjective decisions at category boundaries. Based on these results, we excluded the embedding similarity ranges with a match rate below 0.1 by setting the final thresholds at 0.55 for human--human pairs and 0.45 for human--LLM pairs.

\begin{table}[t]
\centering
\small
\begin{tabular}{lcc}
\toprule
& \multicolumn{2}{c}{Annotated match rate} \\
\cmidrule(lr){2-3}
Stratum & Human--human & Human--LLM \\
\midrule
$< 0.15$       & 0.00 & 0.00 \\
$0.15$--$0.25$ & 0.00 & 0.00 \\
$0.25$--$0.35$ & 0.00 & 0.00 \\
$0.35$--$0.45$ & 0.00 & 0.05 \\
$0.45$--$0.55$ & 0.05 & 0.10 \\
$0.55$--$0.65$ & 0.65 & 0.40 \\
$\geq 0.65$    & 0.75 & 0.65 \\
\bottomrule
\end{tabular}

\caption{Annotated match rate across cosine similarity strata for human--human and human--LLM feedback pairs.}
\label{tab:annotated-match-rate-by-stratum}

\end{table}


\subsection{Automated Feedback Match}
\label{app:fb-assessment-performance}
We automatically match feedback pairs that raise a common concrete point using GPT-5.2. We use the following prompt.
\begin{tcolorbox}[
    colback=gray!5,
    colframe=gray!40,
    fontupper=\ttfamily\small,
    title={\textbf{User Prompt}},
    breakable
]
You are given an abstract of a paper and a pair of feedback items about the paper.

Your task: decide whether the two feedback items match.

\medskip
\textbf{Definition of a match:}
The two feedback include at least one concrete point that shares the following aspects.
\begin{itemize}
\item Targeted part of paper
\item Deficiency pointed out
\item Quality dimension addressed
\item Action requested
\end{itemize}

\medskip
\textbf{Do NOT mark as a match if:}
\begin{itemize}
\item They only refer to the same broad section/topic but raise different issues.
\item One is praise/positive framing while the other is criticism/negative framing (or they otherwise differ in stance).
\item They use similar keywords but the underlying substance or requested change is different.
\end{itemize}

\medskip
\textbf{Paper abstract:} \{abstract\}

\textbf{Feedback 1:} \{feedback1\}

\textbf{Feedback 2:} \{feedback2\}

\medskip
\textbf{Return a JSON object:}
\begin{itemize}
\item \texttt{"match"}: \texttt{"1"} if they match, otherwise \texttt{"0"}
\item \texttt{"explanation"}: a brief justification focusing on the specific overlap (or lack of it)
\end{itemize}

\medskip
\textbf{Output format:}
\begin{verbatim}
[
  {
    "match": "0/1",
    "explanation": "<brief explanation>"
  }
]
\end{verbatim}

\end{tcolorbox}

We use the same human-annotated sample described in Appendix~\ref{app:feedback-pair-filtering} to evaluate the performance of LLM-based feedback match assessment. Sampling feedback pairs uniformly across cosine similarity strata ensures a sufficient number of positive examples in the evaluation set by oversampling the higher-similarity strata relative to their prevalence in the full dataset.

For this evaluation, we retain only pairs whose embedding similarity falls at or above the filtering threshold determined separately for each pair type in the Appendix~\ref{app:feedback-pair-filtering}. We then compare the LLM-generated match labels with the human majority-vote labels and compute accuracy, precision, recall, and F1 score for each stratum (Table~\ref{tab:fb-match-performance-by-stratum}).

Because the evaluation sample is uniformly stratified rather than distributed according to the true similarity distribution of the test set, we additionally report a distribution-weighted F1 score. Specifically, we computed the metrics within each retained stratum and then took their weighted average, using as weight the actual number of feedback pairs in each stratum according to the cosine similarity histogram of the test set of \ours{}-ICLR. This weighting corrects for the sampling imbalance introduced by uniform stratified sampling and yields a summary measure that better reflects performance on the real test set distribution.

\begin{table}[t]
\centering
\small

\begin{subtable}{\linewidth}
\centering
\begin{tabular}{lccccc}
\toprule
Stratum & Distribution weight & Accuracy & Precision & Recall & F1 \\
\midrule
$0.55$--$0.65$ & 0.757 & 0.85 & 1.00 & 0.77 & 0.87 \\
$\geq 0.65$    & 0.243 & 0.80 & 0.92 & 0.80 & 0.86 \\
\midrule
Distribution-weighted & 1.000 & 0.838 & 0.981 & 0.777 & 0.868 \\
\bottomrule
\end{tabular}
\caption*{(a) Human--human feedback pairs}
\end{subtable}

\vspace{0.75em}

\begin{subtable}{\linewidth}
\centering
\begin{tabular}{lccccc}
\toprule
Stratum & Distribution weight & Accuracy & Precision & Recall & F1 \\
\midrule
$0.45$--$0.55$ & 0.753 & 1.00 & 1.00 & 1.00 & 1.00 \\
$0.55$--$0.65$ & 0.210 & 0.70 & 0.67 & 0.50 & 0.57 \\
$\geq 0.65$    & 0.038 & 0.85 & 0.92 & 0.85 & 0.88 \\
\midrule
Distribution-weighted & 1.000 & 0.932 & 0.929 & 0.890 & 0.906 \\
\bottomrule
\end{tabular}
\caption*{(b) Human--LLM feedback pairs}
\end{subtable}

\caption{Performance of LLM-based feedback match assessment across cosine similarity strata. Distribution weights reflect the proportion of feedback pairs in each stratum in the full test set. Accuracy, precision, recall, and F1 are reported only for strata retained after embedding-similarity filtering. The final row in each subtable reports distribution-weighted metrics, computed as weighted averages of stratum-level metrics using the test-set stratum proportions.}
\label{tab:fb-match-performance-by-stratum}
\end{table}

\subsection{Human Consensus-based Feedback Evaluation on Test Set Partitions}
\label{app:cons-eval-results-partition}
\begin{table}[t]
\begin{center}
\small
\setlength{\tabcolsep}{4pt}
\begin{tabular}{lccc|ccc}
\toprule
& \multicolumn{3}{c|}{\bf ICLR 2020--2025} & \multicolumn{3}{c}{\bf ICLR 2026} \\
\bf Model & \bf Precision & \bf Recall & \bf F1 & \bf Precision & \bf Recall & \bf F1 \\
\midrule
Gemini-3-flash    & 0.10 & 0.15 & 0.11 & \textbf{0.15} & \textbf{0.18} & \textbf{0.15} \\
GPT-5.2           & 0.12 & \textbf{0.16} & \textbf{0.12} & 0.14 & 0.17 & 0.14 \\
Llama3.1-8B-Inst. & 0.05 & 0.06 & 0.05 & 0.04 & 0.04 & 0.04 \\
Qwen3-8B (Base)   & 0.08 & 0.10 & 0.08 & 0.06 & 0.07 & 0.06 \\
\midrule
\ours{}-DPO       & 0.07 {\scriptsize(-0.004)} & 0.08 {\scriptsize(-0.019)} & 0.07 {\scriptsize(-0.010)} & 0.11 {\scriptsize(+0.046)} & 0.13 {\scriptsize(+0.058)} & 0.10 {\scriptsize(+0.044)} \\
\ours{}-SFT       & \textbf{0.13} {\scriptsize(+0.051)} & 0.11 {\scriptsize(+0.003)} & 0.10 {\scriptsize(+0.022)} & 0.15 {\scriptsize(+0.083)} & 0.12 {\scriptsize(+0.050)} & 0.11 {\scriptsize(+0.055)} \\
\bottomrule
\end{tabular}
\caption{Performance of \ours{}-trained Qwen3-8B variants and baseline models in matching human consensus feedback on \ours{}-ICLR test sets. For ICLR 2020--2025 ($n=424$), \ours{}-SFT achieves the best performance across all metrics among the size-comparable models (7B/8B). For ICLR 2026 ($n=519$), \ours{}-SFT achieves the best precision and F1 among the size-comparable models, while \ours{}-DPO achieves the best recall. Confidence interval half-widths for all values are less than 0.001.}
\label{tab:result-consensus-fb-eval-partition}
\end{center}
\end{table}

Table~\ref{tab:result-consensus-fb-eval-partition} presents the results of the human consensus-based feedback evaluation separately for ICLR 2020--2025 and ICLR 2026 ($n=424$ and $n=519$, respectively). This analysis helps distinguish performance on earlier papers, which may be more susceptible to contamination, from performance on the temporally held-out 2026 set.
The overall trend is consistent with the aggregated results in the main results. Across both partitions, \ours{} models outperform all size-comparable open-weight baselines, indicating that the gains of our models are not explained solely by possible contamination in the earlier years. 

The relative behavior of the two \ours{} variants shows some differences  across the two partitions. Compared with ICLR 2020--2025, \ours{}-DPO is stronger on ICLR 2026, where it exceeds \ours{}-SFT in recall (0.126 vs.\ 0.118) and narrows the F1 gap (0.102 vs.\ 0.113). This suggests that \ours{}-DPO may be relatively better at generalization in alignment with consensus feedback.

We also observe a change in the relative ranking of the proprietary models across the two partitions. GPT-5.2 performs better than Gemini-3-flash on ICLR 2020--2025, but Gemini-3-flash becomes the strongest on ICLR 2026. Despite this shift, \ours{}-SFT continues to compare favorably in precision: on ICLR 2026, it achieves 0.145 precision, higher than GPT-5.2 (0.141) and close to Gemini-3-flash (0.149). In general, these partitioned results reinforce the robustness of the main findings while highlighting a meaningful temporal variation in model behavior.

\section{Human Evaluation}
\label{app:human-eval}
We recruited a total of $N=13$ authors for human annotations. Of these 13 authors, 35.3\% had 2 years of experience in research and 23.5\% had 6 years of experience with all participants having at least 2 years of research experience. Most participants had research experience in NLP (13), with 2 participants with experience in robotics and ML, and one in HCI. Of the 13 unique papers submitted by participants, 10 of them were submitted to *CL conferences with 2 submitted to ICML and 2 to ICLR. 

\begin{table}[h]
\centering
\small
\caption{Author annotation questions for evaluating LLM-generated feedback.}
\label{tab:annotation-questions}
\begin{tabularx}{\linewidth}{llX}
\toprule
\textbf{\#} & \textbf{Dimension} & \textbf{Question \& Response Options} \\
\midrule
1 & Validity & Do you agree that this feedback is a valid issue/question/suggestion? \\
  &          & \textit{Agree} --- The point is valid. \\
  &          & \textit{Disagree} --- You disagree with the premise, or the reviewer is mistaken. \\
\midrule
2 & Specificity & Is the feedback anchored to specific parts of the paper? (1--5) \\
  &             & 5 = Very specific \quad 4 = Mostly specific \quad 3 = Moderately specific \\
  &             & 2 = Mostly vague \quad\phantom{0} 1 = Very vague \\
\midrule
3 & Action & What action are you willing to take? \\
  &        & \textit{Will revise} --- Make a concrete change to the manuscript. \\
  &        & \textit{Defer to future work} --- Acknowledge but defer as out of scope. \\
  &        & \textit{Point to existing content} --- The paper already addresses this. \\
  &        & \textit{Accept with no revision} --- Valid but no change or deferral. \\
  &        & \textit{Contest with no revision} --- Dispute or reject with no change. \\
  &        & \textit{No action (other)} --- No action for another reason. \\
\midrule
4 & Helpfulness & How useful is the feedback overall to the authors? (1--5) \\
  &             & 5 = Very helpful \quad 4 = Helpful \quad 3 = Moderately helpful \\
  &             & 2 = Slightly helpful \quad 1 = Not helpful \\
\bottomrule
\end{tabularx}
\end{table}

We asked a total of 4 questions for each feedback unit: validity (binary), specificity (Likert scale), author action (select one out of six with an optional text field for details), and helpfulness (Likert scale). See Table~\ref{tab:annotation-questions} for exact wording of questions. 

We randomly sampled 5 feedback units for each model to control for the number of feedback units evaluated per model and to mitigate annotator fatigue. We randomly shuffled the order, and participants saw one feedback unit at a time. 

\section{Baseline Generation}
\label{app:baseline-generation}

We use the following standardized prompt for all baseline models.

\begin{tcolorbox}[
    colback=gray!5,
    colframe=gray!40,
    fontupper=\ttfamily\small,
    title={\textbf{User Prompt}},
    breakable
]
You are an expert researcher. Please review the following research paper and generate a set of constructive feedback to improve it.

\medskip
\textbf{Paper content:} \{paper\_content\}
\end{tcolorbox}

\paragraph{Model Configurations and Infrastructure.}
We use specific API endpoints and snapshots for reproducibility of our results. The open-source models, Llama-3.1-8B-Instruct and Qwen3-8B, are accessed via the Together AI API. For proprietary models, we use GPT-5.2 (\texttt{gpt-5.2-2025-12-11}) through the OpenAI API and Gemini-3-Flash (\texttt{gemini-3-flash-preview}) through the Google Gemini API . 

\paragraph{Decoding and Length Settings.}
We set the decoding temperature to $0.7$ across all models to balance quality and consistency. For Llama-3.1 and Qwen3-8B, the maximum number of new tokens is set to $4096$. In the case of GPT-5.2 and Gemini-3-Flash, where internal reasoning tokens also count toward the total generation limit, we use larger values to accommodate full feedback generation: \texttt{max\_completion\_tokens} is set to $8192$ for GPT-5.2, and \texttt{max\_output\_tokens} is set to its maximum permissible value for Gemini-3-Flash. For Qwen3-8B, we follow the provider's recommended best practices by fixing the sampling parameters to top-p=0.8, top-k=20, and \texttt{min\_p}=0, while other models utilize their respective default settings. In cases where the input prompt exceeds a model's context window, the manuscript is truncated from the end to fit the maximum input token limit.

\section{Aspect Analysis on Novel feedback}
\label{app:novel-feedback}
\begin{table}[t]
\begin{center}
\small
\begin{tabular}{lcccc}
\toprule
\multirow{2}{*}{\bf Aspect} & \multicolumn{2}{c}{\bf \ours{}-SFT} & \multicolumn{2}{c}{\bf \ours{}-DPO} \\
\cmidrule(lr){2-3} \cmidrule(lr){4-5}
 & \bf Novel & \bf Aligned & \bf Novel & \bf Aligned \\
\midrule
Add Experiments on More Datasets & \textbf{0.88} & 0.04 & \textbf{1.73} & 0.56 \\
Add Ablations Experiments        & 0.06 & \textbf{0.38} & \textbf{1.08} & 0.53 \\
Algorithm Efficiency             & -0.30 & \textbf{-0.10} & \textbf{0.08} & 0.05 \\
Theoretical Soundness            & -0.37 & \textbf{0.16} & \textbf{0.14} & 0.09 \\
Implications of the Research     & 0.45 & \textbf{0.81} & \textbf{2.48} & 1.98 \\
Ethical Aspects                  & 1.58 & N/A & 1.49 & N/A \\
Missing Citations                & 11.50 & N/A & 6.85 & N/A \\
Novelty                          & -0.55 & \textbf{-0.18} & -1.00 & -1.00 \\
Clarity and Presentation         & 0.31 & \textbf{0.74} & 0.88 & \textbf{0.96} \\
Comparison to Previous Studies   & \textbf{0.30} & 0.23 & 0.66 & \textbf{1.19} \\
Reproducibility                  & -0.20 & \textbf{0.93} & 0.39 & \textbf{2.66} \\
\midrule
Overall                          & 0.24 & \textbf{0.64} & 0.88 & \textbf{0.93} \\
\bottomrule
\end{tabular}
\end{center}
\caption{Relative improvement over Qwen3-8B in novel and aligned feedback rates by aspect. Both \ours{}-SFT and \ours{}-DPO improve aligned feedback more than novel feedback overall, but \ours{}-DPO shows a more balanced pattern and broader gains across aspects. Bold indicates the larger improvement between novel and aligned feedback within each model. N/A indicates undefined due to zero feedback rate in Qwen3-8B.}
\label{tab:novel-feedback-aspect}
\end{table}
To identify the aspects where gains in the novel feedback rate are most pronounced, we classify each model-generated feedback item using a predefined taxonomy of feedback aspects proposed by \cite{liang2024can}. Each feedback item is annotated with one or more aspects that it emphasizes, when applicable. We integrate this classification into the feedback parsing task, which is performed by gpt-4.1-mini (Appendix~\ref{app:parsing-setup}).
This design follows recent work that has shown that LLM-based classification of aspect in peer review text can produce consistent and useful annotations, validated through consistency analyses and downstream prediction tasks \citet{lu-etal-2025-identifying}.
Table~\ref{tab:novel-feedback-aspect} reports the relative improvement of \ours{} models over Qwen3-8B in both the novel feedback rate (i.e., the proportion of novel feedback items among all generated feedback) and the aligned feedback rate (i.e., the proportion of aligned feedback items among all generated feedback).

\end{document}